\documentclass[runningheads]{llncs}
\usepackage{amsmath}
\usepackage{amssymb}
\usepackage[T1]{fontenc}
\usepackage{multirow}
\usepackage{diagbox}
\usepackage{graphicx}
\usepackage{xcolor}

\usepackage[pagebackref=true,breaklinks=true,colorlinks,bookmarks=false]{hyperref}
\makeatletter
\renewcommand{\thanks}[1]{%
  \@tempcnta\z@
  \footnotetext{#1}%
}
\makeatother
\begin{document}
\title{Two-Stage Mixture-of-LoRA for Multi-Task Medical Vision-Language Learning}
\titlerunning{Two-Stage Mixture-of-LoRA for Medical VLMs}
%
\author{
Zhanghao Chen\textsuperscript{[0009-0000-9249-5124]}\inst{1}\textsuperscript{\#}\thanks{\# Equal Contribution} \and
Yuanyuan Li\textsuperscript{[0009-0003-4469-2926]}\inst{1}\textsuperscript{\#} \and
Zhenyu Lu\textsuperscript{[0009-0004-5954-2004]}\inst{1}\textsuperscript{\#} \and
Shuo Gao\textsuperscript{[0009-0005-8844-4005]}\inst{2} \and
Guangquan Zhou\textsuperscript{[0000-0002-6467-3592]}\inst{2}\textsuperscript{*}\thanks{* Corresponding authors} \and
Yikun Zhang\textsuperscript{[0000-0002-4048-4869]}\inst{1}\textsuperscript{*}
}
\authorrunning{Z. Chen, Y. Li \& Z. Lu et al.}
\institute{%
School of Computer Science and Engineering, Southeast University, Nanjing 211189, China \and
School of Biological Science \& Medical Engineering, Southeast University, Nanjing 211189, China\\
\email{\{guangquan.zhou,yikun\}@seu.edu.cn}
}

\maketitle              
\begin{abstract}
Medical vision-language models (VLMs) allow a single model to perform clinical image analysis tasks ranging from diagnosis classification to report generation. However, joint adaptation is challenged by heterogeneous output formats, conflicting task gradients, and imbalanced training data. Hence, we present \textbf{Two-Stage Mixture-of-LoRA}, a framework built on MedGemma-1.5-4B. The framework uses a shared-specific Mixture-of-LoRA architecture comprising one shared LoRA and six task-specific expert LoRAs, together with a two-stage training procedure. In Stage 1, we jointly train the shared LoRA and all task-specific expert LoRAs on all tasks. In Stage 2, we first freeze the backbone, the shared LoRA, and all non-target experts, and refine one task expert at a time. Classification and regression then receive an additional modality-balanced continuation, in which smaller modality groups are repeated to match the largest group. In the FLARE 2026 Task 3 test sets, the proposed method achieves 0.85 balanced accuracy for classification, 0.48 averaged F1 for multi-label classification, 0.79 detection F1, and 17.40 regression MAE. Code is available \href{https://github.com/YuanYL03/MICCAI-FLARE-2026-Challenge-Task3-2D}{here}.
\end{abstract}
\section{Introduction}
Medical image analysis is a cornerstone of clinical diagnosis, yet traditional task-specific models suffer from limited scalability and high deployment costs in workflows requiring diverse analytical capabilities. Vision-language models (VLMs) enable a unified auto-regressive framework to handle tasks ranging from diagnostic classification to report generation. The FLARE 2026 Task 3 challenge formalizes this setting as a standardized benchmark covering eight imaging modalities and six tasks: classification, multi-label classification, detection, counting, regression, and report generation, while exposing core challenges in adapting a single foundation model to heterogeneous tasks.
The first challenge is \textbf{imbalanced data composition within individual tasks}. Classification combines different imaging modalities, with modality-specific data groups ranging from 64 endoscopy records to 4,206 dental X-ray records; regression contains 5,150 ultrasound records and 702 dental X-ray records. Conventional random shuffling therefore gives larger modality groups more updates, leaving smaller groups under-adapted.
The second challenge is \textbf{heterogeneous and competing task optimization}. Tasks differ drastically in output spaces and reasoning requirements, and empirical results show cross-task trade-offs, e.g., gains in single-label classification often hurt multi-label performance. Forcing all tasks to share one adaptation space leads to conflicting gradients and insufficient specialization.
To address these issues, a holistic solution spanning data sampling, architecture, and optimization strategy is needed. We propose \textbf{Two-Stage Mixture-of-LoRA}, a framework for multi-task medical VLM learning. Our contributions are:
\begin{enumerate}
\item A simple \textbf{within-task sampling scheme} for classification and regression. Within either task, examples from smaller data modality groups are repeated to match the largest modality group.
\item A \textbf{shared-specific Mixture-of-LoRA architecture}. The shared LoRA learns universal representations from all six tasks, whereas each task-specific LoRA learns the visual cues and output format required by its own task.
\item A \textbf{two-stage optimization pipeline} that first jointly learns shared and task-specific knowledge, then independently refines every task expert with the shared LoRA and non-target experts frozen. Classification and regression receive an additional modality-balanced continuation.
\end{enumerate}
\section{Related Work}
\subsection{Medical Vision-Language Models}
General-purpose VLMs such as CLIP~\cite{radford2021clip}, BLIP-2~\cite{li2023blip2}, Qwen2.5-VL~\cite{bai2025qwen25vl}, and Qwen3.5~\cite{qwen3.5} have strong cross-modal capabilities but lack clinical domain knowledge. Domain-specific models such as LLaVA-Med~\cite{li2023llavamed}, HuatuoGPT-Vision~\cite{chen2024towards}, and MedGemma~\cite{sellergren2026medgemma15} address this gap via medical pre-training and instruction tuning. However, most are optimized for open-ended text generation; structured perception tasks such as detection and regression remain underexplored, and few works systematically study multi-task joint-training challenges.
\subsection{Parameter-Efficient Multi-Task Adaptation}
LoRA~\cite{hu2022lora} is a widely used PEFT method for adapting large VLMs efficiently. QLoRA~\cite{dettmers2023qlora} further reduces memory via 4-bit quantization, as applied in ME-VLIP~\cite{shaaban2025mevlip} for medical image parsing. For multi-task settings, Mixture-of-LoRA~\cite{wu2024mixture} uses multiple expert adapters with learned routing, but routing networks add overhead and are unnecessary when task identity is known. Existing Mixture-of-LoRA designs also do not explicitly separate shared and task-specific knowledge, limiting their ability to resolve optimization conflicts.
\subsection{Data Balancing and Optimization in Multi-Task Learning}
Data imbalance and gradient conflicts are longstanding multi-task challenges. Uniform sampling improves under-represented tasks but risks overfitting. Two-stage training alleviates interference by first learning shared representations and then fine-tuning task-specific components. In medical VLMs, existing works typically use default sampling and single-stage training, which are ill-suited for the extreme modality disparity in benchmarks such as FLARE 2026 Task 3. There remains a gap in systematically designing sampling, architecture, and training strategies for multi-task medical vision-language learning.

\section{Method}
\label{sec:method}
\subsection{Preliminary Study}
\label{sec:preliminary}
The challenge considered in this work consists of six heterogeneous medical vision-language tasks, including \textbf{classification, multi-label classification, detection, counting, regression, and report generation.} 
Although these tasks can be unified under an autoregressive vision-language formulation, our preliminary experiments reveal two practical challenges when jointly adapting a single multimodal foundation model to all tasks.

\textbf{Within-task modality imbalance.}
The first challenge arises from imbalanced modality composition within individual tasks. Classification includes modality-specific data groups ranging from 64 endoscopy records to 4,206 dental X-ray records, while regression contains 5,150 ultrasound records and 702 dental X-ray records. Consequently, conventional random shuffling gives the larger modality groups substantially more updates during task-specific training, leaving smaller groups under-adapted. This motivates within-task balanced sampling during expert refinement.

\textbf{Heterogeneous and competing task optimization.}
The second challenge is that different tasks exhibit substantially different optimization behaviors.
Despite sharing the same vision-language backbone, the six tasks differ considerably in their output spaces and reasoning requirements, ranging from short categorical responses to numerical predictions, spatial localization, and long-form report generation.
More importantly, we empirically observe cross-task competition during joint training.
For example, classification and multi-label classification may exhibit a trade-off in which improving one task is accompanied by degraded performance on the other.
Such behavior suggests that forcing all tasks to rely on a single shared adaptation space may lead to conflicting parameter updates and insufficient task specialization.

These observations motivate three components of our framework:
(i) within-task sampling during the additional refinement of classification and regression, giving each modality group the same number of training samples;
(ii) a shared-specific Mixture-of-LoRA architecture that explicitly separates transferable knowledge from task-specific adaptation; and
(iii) a two-stage optimization strategy that first jointly learns shared and
task-specific knowledge, then independently refines every task expert with the
shared LoRA and non-target experts frozen. Classification and regression are
subsequently continued with modality-balanced resampling.

\subsection{Framework Overview}
\label{sec:overview}
\begin{figure}[t]
    \centering
    \includegraphics[width=\textwidth]{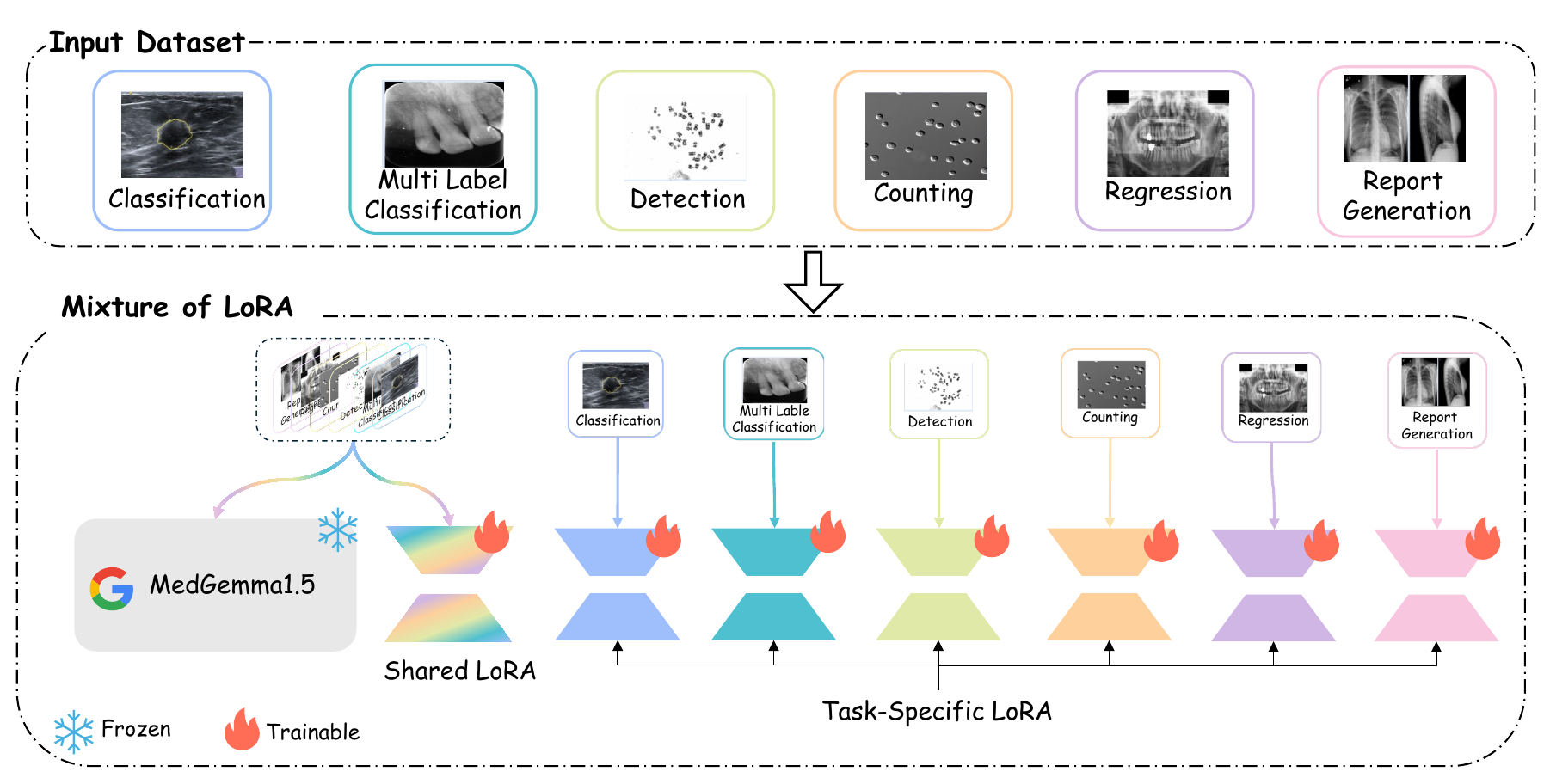}
    \caption{Overview of the proposed Two-Stage Mixture-of-LoRA framework for medical vision-language tasks. The frozen MedGemma-1.5-4B backbone is augmented with one shared LoRA module and six task-specific LoRA experts. For each task, the shared module is jointly activated with its corresponding expert, enabling knowledge sharing across tasks while preserving task-specific adaptation.}
    \label{fig:overview}
\end{figure}

Figure~\ref{fig:overview} illustrates the overall framework.
We adopt the instruction-tuned model MedGemma-1.5-4B~\cite{sellergren2026medgemma15} as the multimodal backbone.
The pretrained MedGemma-1.5 parameters remain frozen throughout training, and only lightweight LoRA modules~\cite{hu2022lora} are optimized.

Let
\[
\mathcal{T}
=
\{
\mathrm{cls},
\mathrm{mlcls},
\mathrm{det},
\mathrm{cnt},
\mathrm{reg},
\mathrm{rep}
\}
\]
denote the set of six tasks.
For each task $t\in\mathcal{T}$, its training dataset is denoted by
\[
\mathcal{D}_t
=
\{
(\mathbf{I}_i,\mathbf{q}_i,\mathbf{y}_i)
\}_{i=1}^{n_t},
\]
where $\mathbf{I}_i$ denotes the medical image, $\mathbf{q}_i$ denotes the corresponding task instruction or question, and $\mathbf{y}_i$ denotes the target response.

We formulate all six tasks as autoregressive text generation.
Different prediction targets, including class labels, multiple labels, bounding-box representations, counts, continuous values, and reports, are converted into task-specific textual responses.
Given a known task identity $t$, the model predicts
\[
p(\mathbf{y}\mid\mathbf{I},\mathbf{q},t)
=
\prod_{m=1}^{|\mathbf{y}|}
p
\left(
y_m
\mid
\mathbf{I},
\mathbf{q},
y_{<m},
t
\right).
\]
Accordingly, all tasks can be optimized using the same autoregressive language modeling objective:
\begin{equation}
\mathcal{L}_{t}
=
-\sum_{m=1}^{|\mathbf{y}|}
\log
p
\left(
y_m
\mid
\mathbf{I},
\mathbf{q},
y_{<m},
t
\right).
\label{eq:sft_loss}
\end{equation}
This unified formulation allows heterogeneous medical tasks to share the same backbone and training objective while relying on task-specific LoRA parameters to capture their distinct prediction behaviors.

\subsection{Mixture-of-LoRA}
\label{sec:molora}
A single LoRA adapter shared by all tasks provides parameter-efficient adaptation but requires the same low-rank parameter space to simultaneously represent heterogeneous task behaviors.
This can be suboptimal when tasks exhibit conflicting optimization directions.
On the other hand, assigning a completely independent adapter to every task prevents beneficial knowledge transfer across related medical tasks.

To balance shared representation learning and task specialization, we introduce a shared-specific Mixture-of-LoRA architecture. The shared LoRA is optimized using samples from all six tasks and captures representations that can be reused across tasks. Each task-specific LoRA is optimized only using its corresponding task and learns the visual cues and output requirements of that task, including class labels, numerical values, bounding boxes, or free-text reports.
For every adapted linear transformation in the frozen backbone, we instantiate one \textbf{shared LoRA} that is accessible to all tasks and \textbf{one task-specific LoRA} for each task.

Consider a frozen linear transformation with weight matrix $\mathbf{W}_0$.
For task $t$, its effective transformation is
\begin{equation}
\mathbf{W}^{(t)}
=
\mathbf{W}_0
+
\Delta\mathbf{W}_{\mathrm{shared}}
+
\Delta\mathbf{W}_{t},
\label{eq:molora}
\end{equation}
where
\begin{equation}
\Delta\mathbf{W}_{\mathrm{shared}}
=
\mathbf{B}_{s}\mathbf{A}_{s},
\qquad
\Delta\mathbf{W}_{t}
=
\mathbf{B}_{t}\mathbf{A}_{t}.
\label{eq:lora_decomposition}
\end{equation}
Here, $\Delta\mathbf{W}_{\mathrm{shared}}$ captures knowledge transferable across tasks, whereas $\Delta\mathbf{W}_{t}$ provides task-specific adaptation capacity.
The standard LoRA scaling factors are omitted from Eq.~\eqref{eq:lora_decomposition} for clarity.

For an input representation $\mathbf{x}$ from task $t$, the corresponding layer therefore computes
\begin{equation}
\mathbf{h}
=
\mathbf{W}_0\mathbf{x}
+
\Delta\mathbf{W}_{\mathrm{shared}}\mathbf{x}
+
\Delta\mathbf{W}_{t}\mathbf{x}.
\label{eq:molora_forward}
\end{equation}

Importantly, our Mixture-of-LoRA does not employ a learned routing network.
Since the task identity is known for each training and inference example, adapter routing is deterministic. The shared LoRA is always activated, while only the task-specific LoRA associated with the current task is selected.
All other task-specific LoRA modules remain inactive for that example.

We insert the shared and task-specific LoRA modules into the same set of backbone transformations.
Specifically, LoRA is applied to the query, key, value, and output projections
(\texttt{q\_proj}, \texttt{k\_proj}, \texttt{v\_proj}, and \texttt{o\_proj})
in the attention modules, as well as the
\texttt{gate\_proj}, \texttt{up\_proj}, and \texttt{down\_proj}
transformations in the feed-forward modules.
The pretrained MedGemma-1.5 parameters remain frozen, making the proposed architecture parameter-efficient while retaining explicit task specialization.

\subsection{Within-Task Sampling}
\label{sec:sampling}
For classification and regression, we divide the training data of each task into predefined modality-specific data groups. Let $\mathcal{G}_h$ denote the groups in task $h$, and let $\mathcal{D}_{h,g}$ be the records in group $g\in\mathcal{G}_h$. Without additional balancing, the number of updates from a group is proportional to $|\mathcal{D}_{h,g}|$, allowing a large group to dominate the refinement of the corresponding expert.

For these tasks, we define $M_h=\max_{g\in\mathcal{G}_h}|\mathcal{D}_{h,g}|$ and construct a resampled continuation set
\begin{equation}
\widetilde{\mathcal{D}}_h=\bigcup_{g\in\mathcal{G}_h}\operatorname{Repeat}(\mathcal{D}_{h,g},M_h),
\label{eq:within_task_sampling}
\end{equation}
where $\operatorname{Repeat}(\cdot,M_h)$ contains complete copies of a group plus a seeded shuffled remainder, giving exactly $M_h$ records for every group. The resulting records are globally shuffled with a fixed seed and materialized as a training file. Thus, each group contributes equally in an epoch. Sampling is performed within an individual task: it neither mixes tasks nor makes the total sample counts of different tasks equal. If a task contains only one group, its data remain unchanged. This is distinct from a stochastic weighted sampler and makes the effective data distribution reproducible.

\subsection{Two-Stage Training}
\label{sec:two_stage}
\begin{figure}[h]
    \centering
    \includegraphics[width=\textwidth]{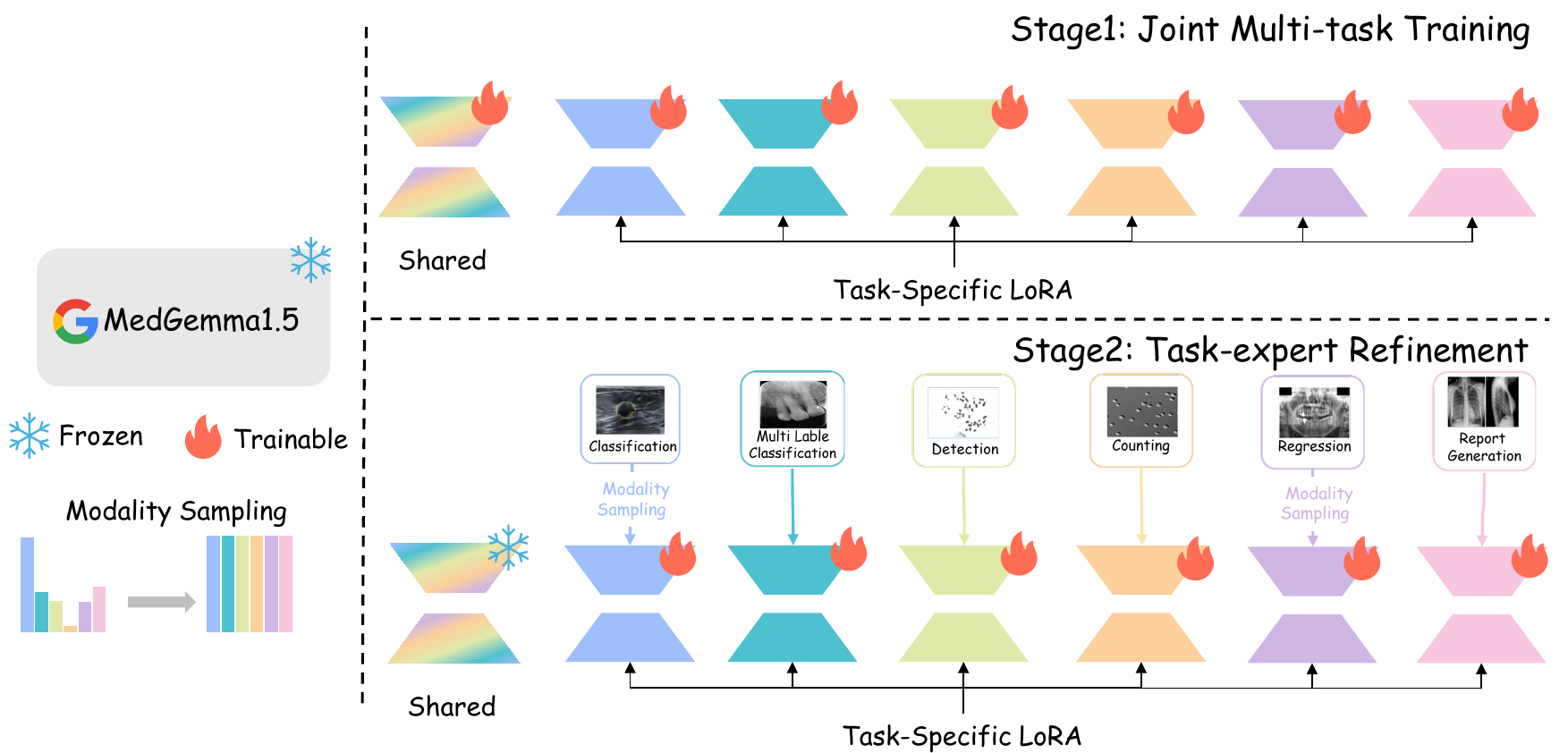}
    \caption{Two-stage optimization strategy.
    In Stage 1, the shared LoRA and all task-specific LoRAs are jointly trained on the six tasks while the MedGemma-1.5 backbone remains frozen.
    In Stage 2, all six experts are independently refined while the backbone, shared LoRA, and non-target experts remain frozen. Classification and regression then receive an additional continuation using modality-balanced resampling.}
    \label{fig:two_stage}
\end{figure}

Although task-specific adapters alleviate parameter-level competition among heterogeneous tasks, different tasks may still exhibit different convergence rates.
A fixed joint-training schedule can therefore terminate before some tasks are sufficiently optimized, while continuing joint optimization for substantially longer may degrade tasks that have already converged.
We address this issue using the two-stage optimization strategy illustrated in Figure~\ref{fig:two_stage}.

\textbf{Stage 1: Joint multi-task training.}
In the first stage, the shared LoRA and all six task-specific LoRAs are jointly optimized on all training tasks. The pretrained MedGemma-1.5 backbone remains frozen, whereas both the shared LoRA parameters and all task-specific LoRA parameters are trainable.

Let $\boldsymbol{\phi}_{s}$ denote the parameters of the shared LoRA and
$\boldsymbol{\phi}_{t}$ denote the task-specific LoRA parameters of task $t$. Let $\mathcal{D}_{\mathrm{all}}=\bigcup_{t\in\mathcal{T}}\mathcal{D}_t$ denote the union of all task-specific training sets.
The Stage~1 objective can be written as
\begin{equation}
\min_{\boldsymbol{\phi}_{s},
\{\boldsymbol{\phi}_{t}\}_{t\in\mathcal{T}}}
\mathbb{E}_{(\mathbf{I},\mathbf{q},\mathbf{y})\sim{\mathcal{D}_{\mathrm{all}}}}
\left[
\mathcal{L}_{t}
\right].
\label{eq:stage1}
\end{equation}
This stage encourages the shared LoRA to capture basic transferable representations across heterogeneous medical tasks, while each task-specific LoRA simultaneously learns initial adaptation for its corresponding task.

\textbf{Stage 2: Independent expert refinement.}
After the five epochs of Stage~1, every task expert is independently refined from the same Stage~1 endpoint. For each task $t\in\mathcal{T}$, we freeze the MedGemma-1.5 backbone, the shared LoRA, and all task-specific LoRAs except $\boldsymbol{\phi}_{t}$. Thus, only the expert associated with task $t$ is updated, using its original task-specific training data $\mathcal{D}_t$. The initial refinement objective is
\begin{equation}
\boldsymbol{\phi}_{t}^{\mathrm{ref}}
=
\arg\min_{\boldsymbol{\phi}_{t}}
\mathbb{E}_{(\mathbf{I},\mathbf{q},\mathbf{y})\sim\mathcal{D}_t}
\left[
\mathcal{L}_{t}
\right].
\label{eq:stage2}
\end{equation}

For classification and regression, we further continue from $\boldsymbol{\phi}_{t}^{\mathrm{ref}}$ for three epochs using the resampled set $\widetilde{\mathcal{D}}_t$ from Eq.~\eqref{eq:within_task_sampling}. The same parameters remain frozen, and only $\boldsymbol{\phi}_{t}$ is updated:
\begin{equation}
\boldsymbol{\phi}_{t}^{*}
=
\arg\min_{\boldsymbol{\phi}_{t}}
\mathbb{E}_{(\mathbf{I},\mathbf{q},\mathbf{y})\sim\widetilde{\mathcal{D}}_t}
\left[
\mathcal{L}_{t}
\right],
\quad t\in\{\mathrm{cls},\mathrm{reg}\},
\label{eq:stage2_balanced}
\end{equation}
where the optimization is initialized from $\boldsymbol{\phi}_{t}^{\mathrm{ref}}$. For the other four tasks, we set $\boldsymbol{\phi}_{t}^{*}=\boldsymbol{\phi}_{t}^{\mathrm{ref}}$. This additional continuation gives every modality-specific group equal exposure in an epoch. The other four tasks use only the initial refinement on their original task-specific training data.

Freezing the shared LoRA preserves the task-general knowledge acquired during Stage 1, and freezing non-target experts prevents refinement of one task from modifying the adapters of other tasks. At inference time, a task activates the shared LoRA learned in Stage 1 together with its corresponding task-specific adapter. We select the final checkpoint independently for each task using the available validation evidence. This design preserves the benefits of joint multi-task learning while allowing every expert to adapt independently.

\newcommand{\PVH}[2]{%
    \makebox[3.8em][r]{#1}\,|\,\makebox[3.8em][l]{#2}%
}

\section{Experiments}

\subsection{Dataset and evaluation measures}

We conduct all experiments on the official FLARE 2026 Task 3-2D dataset, which covers six tasks: disease diagnosis classification, multi-label classification, report generation, lesion detection, cell counting, and numerical regression. The training, public-validation, and hidden-validation splits contain 44,639, 5,577, and 1,783 examples, respectively. In Tables~\ref{tab:comparison} and~\ref{tab:ablation}, each entry is reported as public validation $|$ hidden validation, and NR indicates that the corresponding tasks on validation-hidden dataset are not reported by organizers.

\begin{figure*}[t]
    \centering
    \includegraphics[width=0.72\textwidth]{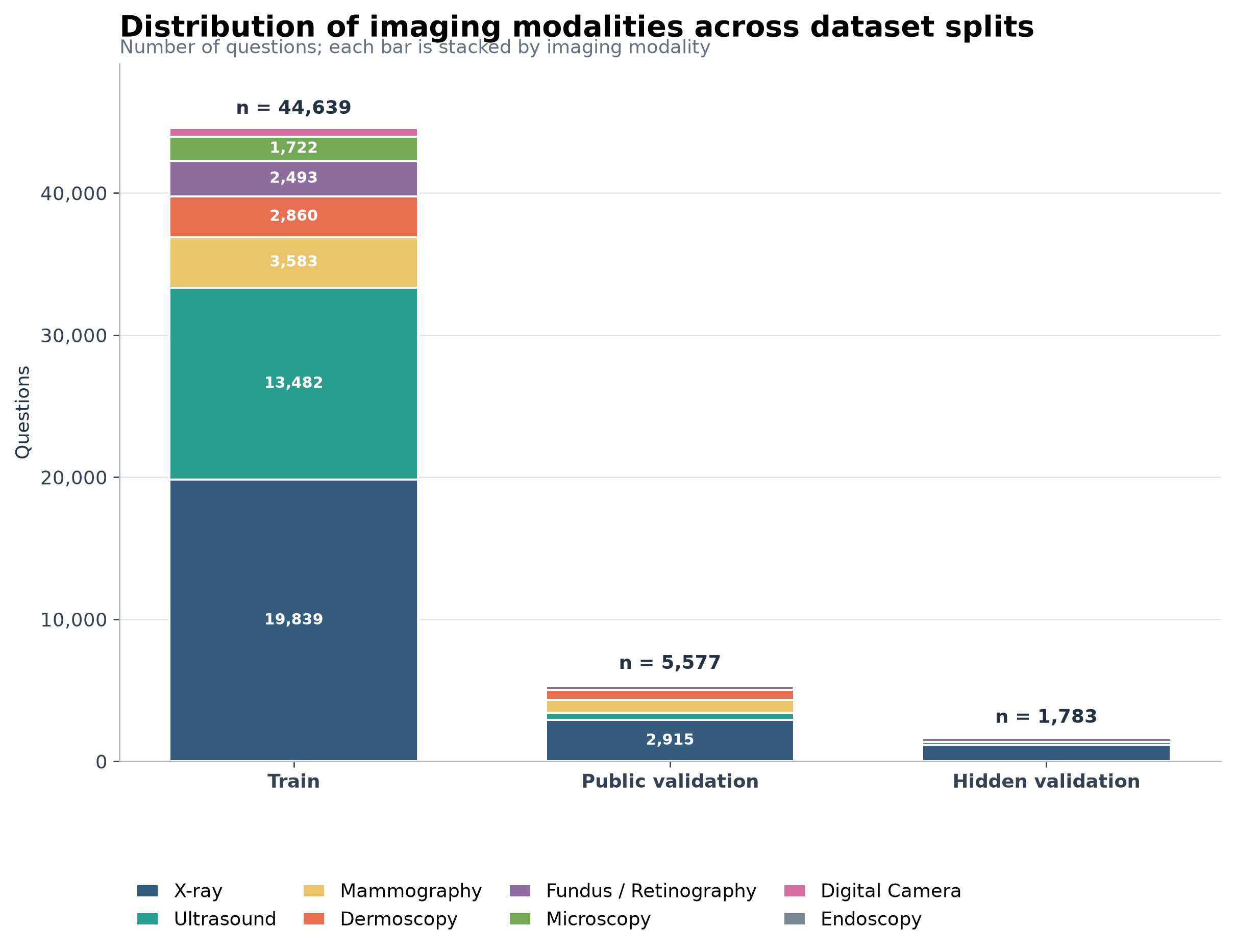}
    \caption{Distribution of imaging modalities across the training, public-validation, and hidden-validation splits. Each bar is stacked by modality and labeled with its total number of questions.}
    \label{fig:modality_by_split}
\end{figure*}

\begin{figure*}[t]
    \centering
    \includegraphics[width=\textwidth]{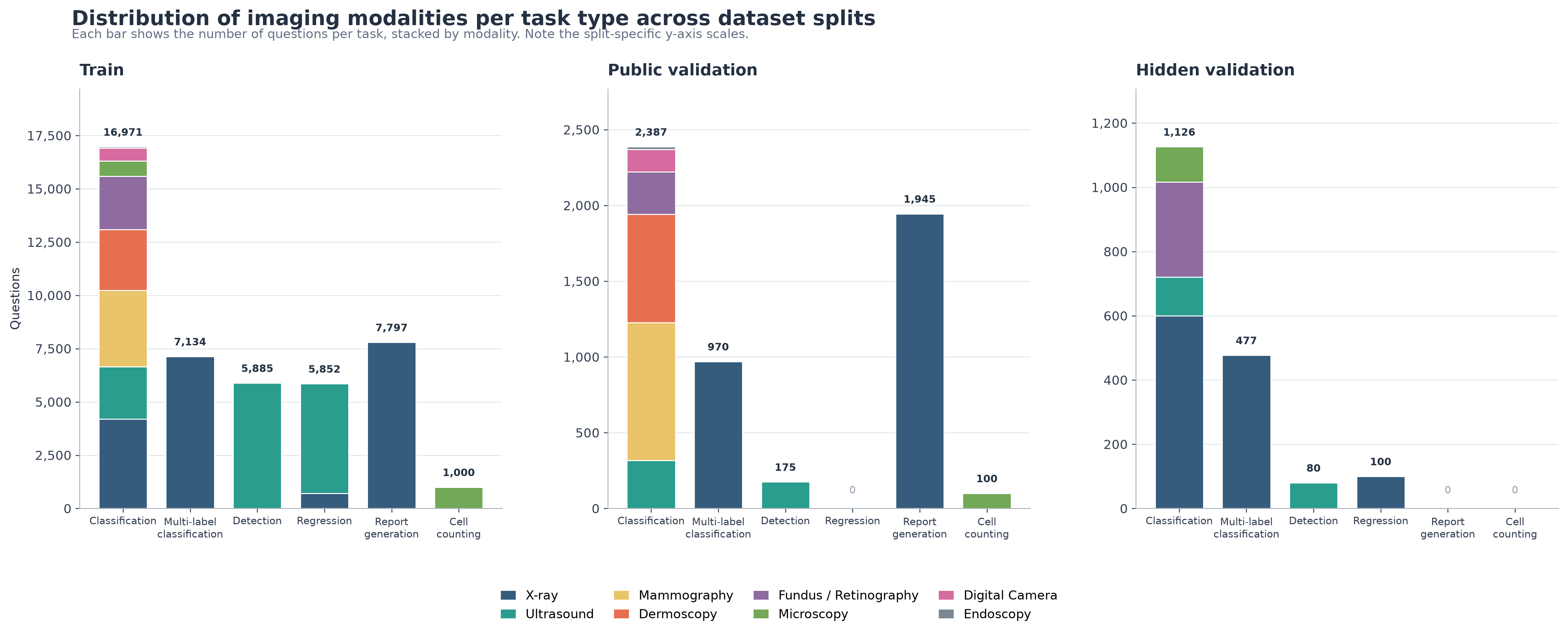}
    \caption{Task-wise modality distribution across the training, public-validation, and hidden-validation splits. The three panels use split-specific vertical scales.}
    \label{fig:modality_by_task_split}
\end{figure*}

Figures~\ref{fig:modality_by_split} and~\ref{fig:modality_by_task_split} summarize the composition of the three splits by imaging modality and task. The modality distribution differs across splits, and the task-level breakdown is also split dependent: regression is absent from public validation, whereas report generation and cell counting are absent from hidden validation. These differences make task-wise evaluation and checkpoint selection necessary rather than relying on a single aggregate score. 

We apply EXIF orientation correction, convert each image to RGB, and resize it to \(896\times896\) using bicubic interpolation.
The image processor rescales pixel values to \([0,1]\) and normalizes each channel with mean and standard deviation of 0.5.
We do not apply padding or stochastic image augmentation.
At most one image is used per example.
All tasks use the model's standard chat template and tokenizer, without special tokens for coordinates or numbers.
For detection, each box is represented as text in the original image's absolute pixel coordinates, \([x_{\mathrm{min}}, y_{\mathrm{min}}, x_{\mathrm{max}}, y_{\mathrm{max}}]\). When an annotation contains multiple boxes, their bracketed coordinate lists are separated by semicolons. Although the input image is resized to \(896\times896\), the detection targets and model outputs use original-image pixel coordinates directly; no coordinate conversion is applied. On the public-validation set, all 175 detection outputs were parseable by our local parser (0/175 unparseable). Training loss is applied to the assistant response rather than the image and prompt tokens.

We use the official metrics for each task. Disease diagnosis classification is evaluated by balanced accuracy, which averages the recall of all classes and is less sensitive to class imbalance. Multi-label classification uses macro F1, computed from the global numbers of true positives, false positives, and false negatives. Detection uses F1 at an intersection-over-union (IoU) threshold of 0.5. Report generation is evaluated with CRIMSON~\cite{baharoon2026crimson}, while counting and regression use MAE, for which lower values indicate better performance.

\subsection{Implementation details}

\subsubsection{Environment settings}

We use MedGemma-1.5-4B as the multimodal backbone and perform parameter-efficient adaptation with LoRA. We use MedGemma because it is a medical-domain vision-language model with strong prior knowledge of radiology and clinical language. Its efficient LoRA fine-tuning also makes it practical to adapt one shared backbone to our diverse medical image tasks. The pretrained backbone remains frozen throughout training, while LoRA modules are inserted into the specified attention and MLP projections. The resulting model contains one shared LoRA adapter and six task-specific experts. Given the task identity, the shared adapter and the corresponding task expert are activated jointly during inference.

\begin{table}[!htbp]
\caption{Development environments and requirements.}
\label{table:env}
\centering
\begin{tabular}{ll}
\hline
System & Ubuntu 22.04.3 LTS \\
\hline
CPU & Intel Xeon Platinum 8488C \\
\hline
GPU & 2 $\times$ NVIDIA RTX PRO 6000 Blackwell (96 GiB each) \\
\hline
CUDA version & 13.0 \\
\hline
Programming language & Python 3.14.6 \\
\hline
Deep learning framework & PyTorch 2.13.0; Transformers 5.14.1 \\
\hline
Specific dependencies & PEFT 0.19.1; Accelerate 1.14.0; TRL 1.8.0 \\
\hline

\end{tabular}
\end{table}

Experiments are conducted on two NVIDIA RTX PRO 6000 Blackwell GPUs, each with 96 GiB of GPU memory. The development environment is summarized in Table~\ref{table:env}, where TRL stands for Transformer Reinforcement Learning and PEFT denotes Parameter-Efficient Fine-Tuning.

\begin{table}[!htbp]
\centering
\caption{Task-specific refinement epochs for Stage 2 training.}
\label{tab:stage2_epochs}
\begin{tabular}{l|c}
\hline
\textbf{Task Name} & \textbf{Refinement Epochs} \\ \hline
Disease Diagnosis Classification & 3 \\
Multi-Label Classification & 5 \\
Lesion Detection & 8 \\
Cell Counting & 3 \\
Numerical Regression & 8 \\
Report Generation & 3 \\ \hline
\end{tabular}
\end{table}

\subsubsection{Training protocols}

Our training procedure consists of two stages. In Stage~1, the shared LoRA adapter and six task-specific experts are jointly optimized for 5 epochs using all training instances. In Stage~2, each task expert is independently refined from the Stage~1 endpoint with the backbone, shared LoRA, and non-target experts frozen. These task-specific epoch counts are summarized in Table~\ref{tab:stage2_epochs}. The specific refinement epoch counts for each task were determined via 5-fold cross-validation solely on the training set. Classification and regression then receive an additional 3-epoch continuation using modality-balanced resampling.

The core training and inference settings are summarized in Table~\ref{table:training}. For the Stage~1 ablation configurations in Table~\ref{tab:ablation}, Shared LoRA has 38.50M trainable LoRA parameters in total and activates 38.50M per example. Expert LoRA has 230.99M trainable LoRA parameters across six experts and activates 38.50M per example. Mixture-of-LoRA has 269.48M trainable LoRA parameters in total and activates 77.00M per example, comprising the shared adapter and the task-specific expert. For multi-image inputs, we use at most the first image of the sample.

\begin{table}[!htbp]
\caption{Core training and inference settings.}
\label{table:training}
\centering
\begin{tabular}{ll}
\hline
Backbone & MedGemma-1.5-4B \\
\hline
LoRA & Rank 16, $\alpha=16$ \\
\hline
Input resolution & $896\times896$ \\
\hline
Precision & BF16 \\
\hline
Batch size & 16 (global) \\
\hline
Optimizer & Fused AdamW \\
\hline
Learning rate (LR) & $2\times10^{-4}$ \\
\hline
LR schedule & Cosine annealing with linear warmup \\
\hline
Training strategy & Two-stage task-specific refinement \\
\hline
Inference & Greedy decoding ($T=0$) \\
\hline
Training time & About 24 hours \\
\hline
End-to-end total time & 80.62 s \\
\hline
End-to-end latency per example & 1.34 s/example \\
\hline
Peak GPU memory of the inference process & About 10.50 GiB \\
\hline
Image size reported by Docker CLI & 30.5 GB \\
\hline
Compressed submission package size & 12.26 GB \\
\hline
Stage-1 trainable LoRA parameters & 269.48M \\
\hline
Stage-2 trainable parameters per run & 38.50M \\
\hline
Total trainable parameters & About 308M \\
\hline
LoRA parameters activated per inference example & 77.00M \\
\hline
\end{tabular}
\end{table}

\subsection{Compared methods}

We compare our approach with Qwen2.5-VL-7B~\cite{bai2025qwen25vl}, Hulu-Med-7B~\cite{jiang2025hulu}, Lingshu-7B~\cite{xu2026lingshu}, InternVL3-8B~\cite{zhu2025internvl3}, Qwen3.5-9B~\cite{qwen3.5}, Qwen3-VL-4B~\cite{bai2025qwen3}, and vanilla MedGemma-1.5-4B~\cite{sellergren2026medgemma15}. Fairly, these baseline models are fine-tuned using a single LoRA with same hyper-parameters, covering both general-purpose and medical-domain VLMs and provide a broad comparison across the six tasks.

For our final MedGemma-based model, the reported results are obtained through task-wise checkpoint selection rather than a single common refinement checkpoint. For tasks with public-validation labels, we select the best-performing saved checkpoint according to the corresponding public-validation metric. Regression is absent from public validation. For this task, we use the endpoint of the modality-balanced continuation.

\section{Results and discussion}

\subsection{Quantitative results on validation set}

\subsubsection{Overall comparison}

Table~\ref{tab:comparison} summarizes the performance of representative medical and general-purpose VLMs on the FLARE 2026 Task 3-2D dataset.

\begin{table*}[!htbp]
\caption{Comparison with medical VLM baselines on FLARE-MLLM-2D.
Values are public validation $|$ hidden validation.
$\uparrow$ indicates higher is better, while $\downarrow$ indicates lower is better.}
\label{tab:comparison}
\centering

\resizebox{\textwidth}{!}{%
\begin{tabular}{lccc}
\hline
\multirow{2}{*}{\diagbox{Method}{Task}}
& \textbf{Classification}
& \textbf{Detection}
& \textbf{Multi-label}
\\

&
Balanced Acc. $\uparrow$
& F1 $\uparrow$
& Micro-F1 $\uparrow$
\\
\hline

Qwen2.5-VL-7B
& \PVH{0.3600}{0.5100}
& \PVH{\textbf{0.6400}}{N/A}
& \PVH{0.3600}{0.5300}
\\

Hulu-Med-7B
& \PVH{0.4413}{0.6353}
& \PVH{0.3666}{N/A}
& \PVH{0.5523}{0.4632}
\\
\hline

Lingshu-7B
& \PVH{0.5203}{0.6528}
& \PVH{0.3048}{N/A}
& \PVH{0.5295}{0.4756}
\\
\hline

InternVL3-8B
& \PVH{0.5200}{0.7100}
& \PVH{0.3700}{N/A}
& \PVH{0.4600}{\textbf{0.5700}}
\\
\hline

Qwen3.5-9B
& \PVH{0.5374}{0.7161}
& \PVH{0.6207}{N/A}
& \PVH{0.5947}{0.4824}
\\
\hline

Qwen3-VL-4B
& \PVH{0.5137}{0.7200}
& \PVH{0.3904}{N/A}
& \PVH{0.5661}{0.4800}
\\
\hline

MedGemma-1.5-4B
& \PVH{0.5547}{0.7831}
& \PVH{0.2005}{N/A}
& \PVH{\textbf{0.6075}}{0.4449}
\\
\hline

Ours
& \PVH{\textbf{0.5656}}{\textbf{0.8694}}
& \PVH{0.2727}{N/A}
& \PVH{0.5942}{0.4917}
\\

\hline
\end{tabular}
}

\vspace{0.15cm}

\resizebox{\textwidth}{!}{%
\begin{tabular}{lccc}
\hline
\multirow{2}{*}{\diagbox{Method}{Task}}
& \textbf{Report Generation}
& \textbf{Regression}
& \textbf{Counting}
\\

&
CRIMSON $\uparrow$
& MAE $\downarrow$
& MAE $\downarrow$
\\
\hline

Qwen2.5-VL-7B
& \PVH{0.8300}{N/A}
& \PVH{N/A}{15.7000}
& \PVH{243.60}{N/A}
\\
\hline

Hulu-Med-7B
& \PVH{0.3603}{N/A}
& \PVH{N/A}{13.8372}
& \PVH{250.04}{N/A}
\\
\hline

Lingshu-7B
& \PVH{0.8525}{N/A}
& \PVH{N/A}{\textbf{12.6588}}
& \PVH{222.55}{N/A}
\\
\hline

InternVL3-8B
& \PVH{0.7500}{N/A}
& \PVH{N/A}{18.6700}
& \PVH{301.40}{N/A}
\\
\hline

Qwen3.5-9B
& \PVH{0.8762}{N/A}
& \PVH{N/A}{16.2751}
& \PVH{\textbf{172.68}}{N/A}
\\
\hline

Qwen3-VL-4B
& \PVH{\textbf{0.8765}}{N/A}
& \PVH{N/A}{18.2900}
& \PVH{262.06}{N/A}
\\
\hline

MedGemma-1.5-4B
& \PVH{0.8561}{N/A}
& \PVH{N/A}{15.5031}
& \PVH{275.64}{N/A}
\\
\hline

Ours
& \PVH{0.8678}{N/A}
& \PVH{N/A}{12.8159}
& \PVH{255.93}{N/A}
\\

\hline
\end{tabular}
}

\end{table*}

For classification, our model achieves a hidden balanced accuracy of 0.8694, improving over vanilla MedGemma from 0.7831 by 0.0863 and obtaining the best reported result. For regression, our hidden MAE decreases from 15.5031 to 12.8159, corresponding to a 17.3\% reduction and ranking second among the reported methods.

The proposed expert-wise refinement improves detection and counting over the vanilla MedGemma baseline, increasing detection F1 from 0.2005 to 0.2727 and reducing counting MAE from 275.64 to 255.93. Meanwhile, our model remains competitive on multi-label classification and report generation.

Among the external baselines, Qwen2.5-VL-7B achieves a higher detection F1 of 0.6400, while InternVL3-8B obtains the highest reported hidden multi-label F1 of 0.5700. Thus, our method does not lead on every task, but it provides competitive performance across the benchmark and is particularly strong on disease diagnosis classification, where it attains the highest hidden balanced accuracy of 0.8694.

\subsubsection{Ablation study}

Table~\ref{tab:ablation} separates the effects of the Stage~1 adapter design, expert-wise refinement, and modality-balanced
resampling. The first three rows compare Shared LoRA, Expert LoRA (trains six task-routed rank-16 adapters for five epochs with others remaining unchanged), and Mixture-of-LoRA. The Expert-wise Refinement row updates one task expert at a time from the Mixture-of-LoRA checkpoint. The Modality-balanced Resampling row reports the final
task-wise checkpoints, with additional resampling runs for classification and regression.

The Stage~1 adapter configurations show different strengths across tasks. Expert LoRA achieves the highest public classification score (0.5976) and report-generation CRIMSON score (0.8704). Shared LoRA performs best on public multi-label classification (F1 = 0.6075) and achieves a detection F1 of 0.2005, substantially exceeding Expert LoRA (0.1141), although it is less competitive on some other tasks. With the same five-epoch training budget, Shared LoRA requires fewer GPU hours but is limited by the lack of task-specific specialization. Mixture-of-LoRA achieves the best hidden classification balanced accuracy at 0.8683 and the lowest hidden regression MAE at 12.9630 among the Stage~1 variants. Its detection F1 is 0.1340, and it provides the checkpoint used for subsequent refinement. Compared with Shared LoRA, Mixture-of-LoRA has lower public-validation detection F1 (0.1340 vs. 0.2005) and multi-label F1 (0.6027 vs. 0.6075), suggesting possible negative transfer during Stage~1 joint training.

Compared with Mixture-of-LoRA, expert-wise refinement leaves classification unchanged at 0.5651 $|$ 0.8683, but increases detection
F1 from 0.1340 to 0.2727 (The two public validation sub-datasets, BUS-UCLM and BUSI, 0.2917 and 0.2662 for each) and hidden multi-label F1 from 0.4214 to 0.4917. It also improves report-generation CRIMSON from 0.8633
to 0.8678, reduces regression MAE from 12.9630 to 12.8853, and reduces counting MAE from 279.06 to 255.93. The public multi-label
score decreases from 0.6027 to 0.5942.

Modality-balanced resampling provides additional gains for the two targeted tasks. Classification reaches 0.5656 $|$ 0.8694 on the
public and hidden splits, respectively, while regression MAE decreases from 12.8853 to 12.8159. The other tasks retain their
expert-wise refinement results.

\begin{table*}[!htbp]
\caption{Ablation of the proposed two-stage adaptation strategy (Shared LoRA and Expert LoRA are alternative designs, others build on the row above).
Values are public validation $|$ hidden validation.
$\uparrow$ indicates higher is better, while $\downarrow$ indicates lower is better.}
\label{tab:ablation}
\centering

\resizebox{\textwidth}{!}{%
\begin{tabular}{lccc}
\hline

\multirow{2}{*}{\diagbox{Method}{Task}}
& \textbf{Classification}
& \textbf{Detection}
& \textbf{Multi-label}
\\

&
Balanced Acc. $\uparrow$
& F1 $\uparrow$
& Micro-F1 $\uparrow$
\\
\hline

Shared LoRA
& \PVH{0.5547}{0.7831}
& \PVH{0.2005}{N/A}
& \PVH{\textbf{0.6075}}{0.4449}
\\ \hline

Expert LoRA
& \PVH{\textbf{0.5976}}{0.7571}
& \PVH{0.1141}{N/A}
& \PVH{0.5885}{0.4484}
\\ \hline

Mixture-of-LoRA
& \PVH{0.5651}{0.8683}
& \PVH{0.1340}{N/A}
& \PVH{0.6027}{0.4214}
\\ \hline

Expert-wise Refinement
& \PVH{0.5651}{0.8683}
& \PVH{\textbf{0.2727}}{N/A}
& \PVH{0.5942}{\textbf{0.4917}}
\\

\hline

Modality-balanced Resampling
& \PVH{0.5656}{\textbf{0.8694}}
& \PVH{\textbf{0.2727}}{N/A}
& \PVH{0.5942}{\textbf{0.4917}}
\\

\hline
\end{tabular}
}

\vspace{0.15cm}

\resizebox{\textwidth}{!}{%
\begin{tabular}{lccc}
\hline

\multirow{2}{*}{\diagbox{Method}{Task}}
& \textbf{Report Generation}
& \textbf{Regression}
& \textbf{Counting}
\\

&
CRIMSON $\uparrow$
& MAE $\downarrow$
& MAE $\downarrow$
\\
\hline

Shared LoRA
& \PVH{0.8561}{N/A}
& \PVH{N/A}{15.5031}
& \PVH{275.64}{N/A}
\\ \hline

Expert LoRA
& \PVH{\textbf{0.8704}}{N/A}
& \PVH{N/A}{15.0879}
& \PVH{286.98}{N/A}
\\ \hline

Mixture-of-LoRA
& \PVH{0.8633}{N/A}
& \PVH{N/A}{12.9630}
& \PVH{279.06}{N/A}
\\ \hline

Expert-wise Refinement
& \PVH{0.8678}{N/A}
& \PVH{N/A}{12.8853}
& \PVH{\textbf{255.93}}{N/A}
\\

\hline

Modality-balanced Resampling
& \PVH{0.8678}{N/A}
& \PVH{N/A}{\textbf{12.8159}}
& \PVH{\textbf{255.93}}{N/A}
\\

\hline
\end{tabular}
}

\end{table*}

\subsubsection{Task-wise analysis}

\begin{figure*}[!htbp]
    \centering
    \includegraphics[width=0.99\linewidth]{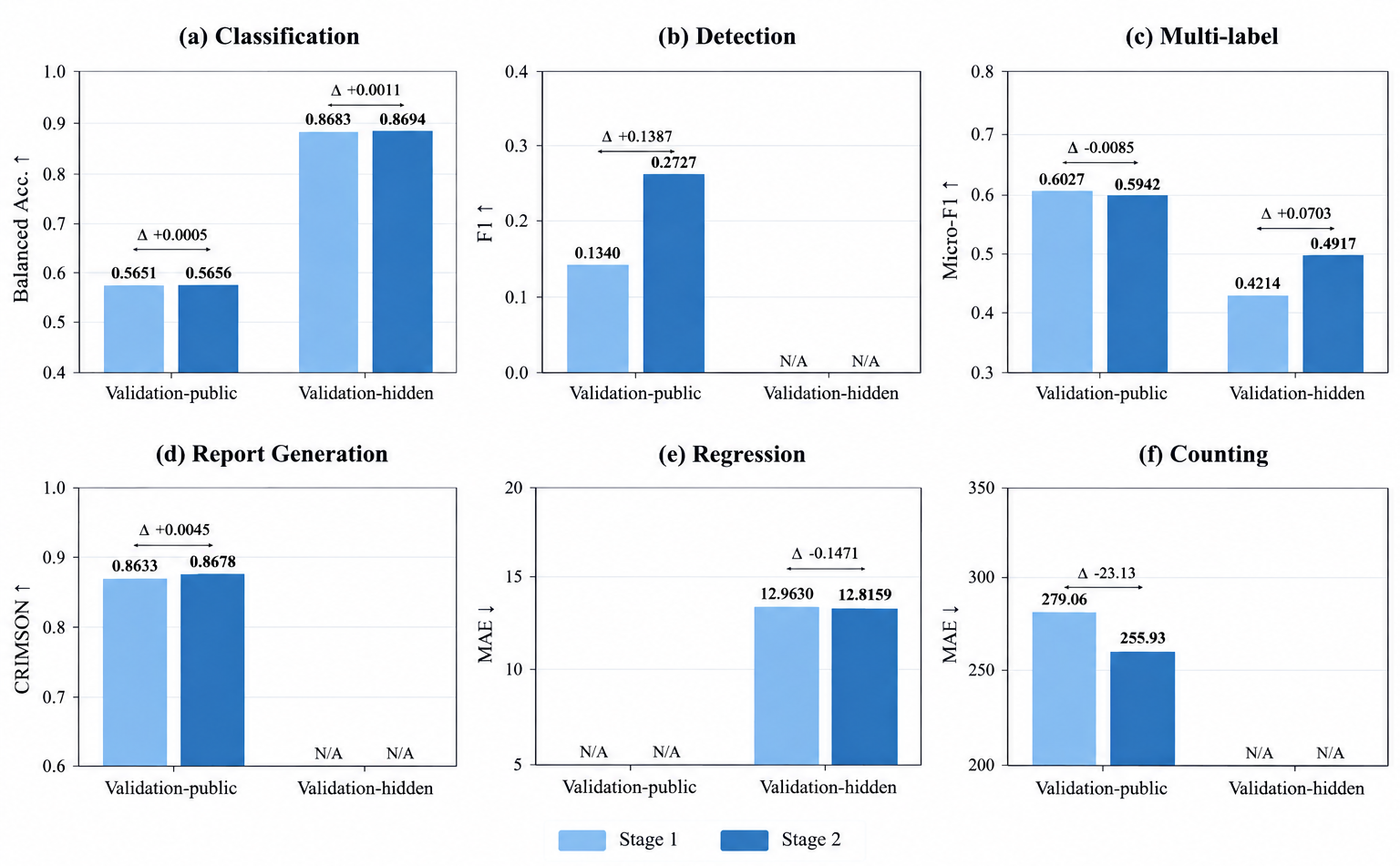}
    \caption{Changes in validation performance from Mixture-of-LoRA to expert-wise refinement and modality-balanced resampling. Expert-wise refinement improves detection, counting, report generation, hidden multi-label classification, and regression, while public multi-label performance decreases slightly. Modality-balanced resampling provides additional gains for classification and regression (Table \ref{tab:ablation}).}
    \label{fig:taskwise}
\end{figure*}

Figure~\ref{fig:taskwise} summarizes the effects of the two Stage~2 components. Expert-wise refinement increases detection F1 by
0.1387, decreases counting MAE by 23.13, raises hidden multi-label F1 by 0.0703, improves report-generation CRIMSON by 0.0045,
and reduces regression MAE by 0.0777 (Table \ref{tab:ablation}); public multi-label F1 decreases by 0.0085. Modality-balanced resampling further reduces
regression MAE by 0.0694 (Table \ref{tab:ablation}). Relative to Mixture-of-LoRA, the classification resampling branch improves public and hidden balanced
Accuracy by 0.0005 and 0.0011, respectively.

\subsubsection{Classification performance by modality}

Table~\ref{tab:classification_modality} reports classification performance by imaging modality on the public-validation set, including the number of samples, balanced accuracy, and accuracy. Regression is not included because no regression samples are available in the public-validation set.

\begin{table}[!htbp]
\centering
\caption{Classification performance by imaging modality on the public-validation set. Balanced accuracy and accuracy are reported per modality; higher values are better.}
\label{tab:classification_modality}
\begin{tabular}{lrrr}
\hline
Modality & Samples & Balanced Acc. & Accuracy \\
\hline
Dermoscopy & 716 & 0.4780 & 0.6578 \\
Digital Camera & 149 & 0.7716 & 0.7852 \\
Endoscopy & 16 & 0.2714 & 0.4375 \\
Mammography & 910 & 0.5045 & 0.7055 \\
Retinography & 279 & 0.7673 & 0.8280 \\
Ultrasound & 317 & 0.8908 & 0.9054 \\
\hline
\end{tabular}
\end{table}

\subsection{Qualitative results on validation set}

\begin{figure*}[!htbp]
  \centering
  \includegraphics[width=0.99\textwidth]{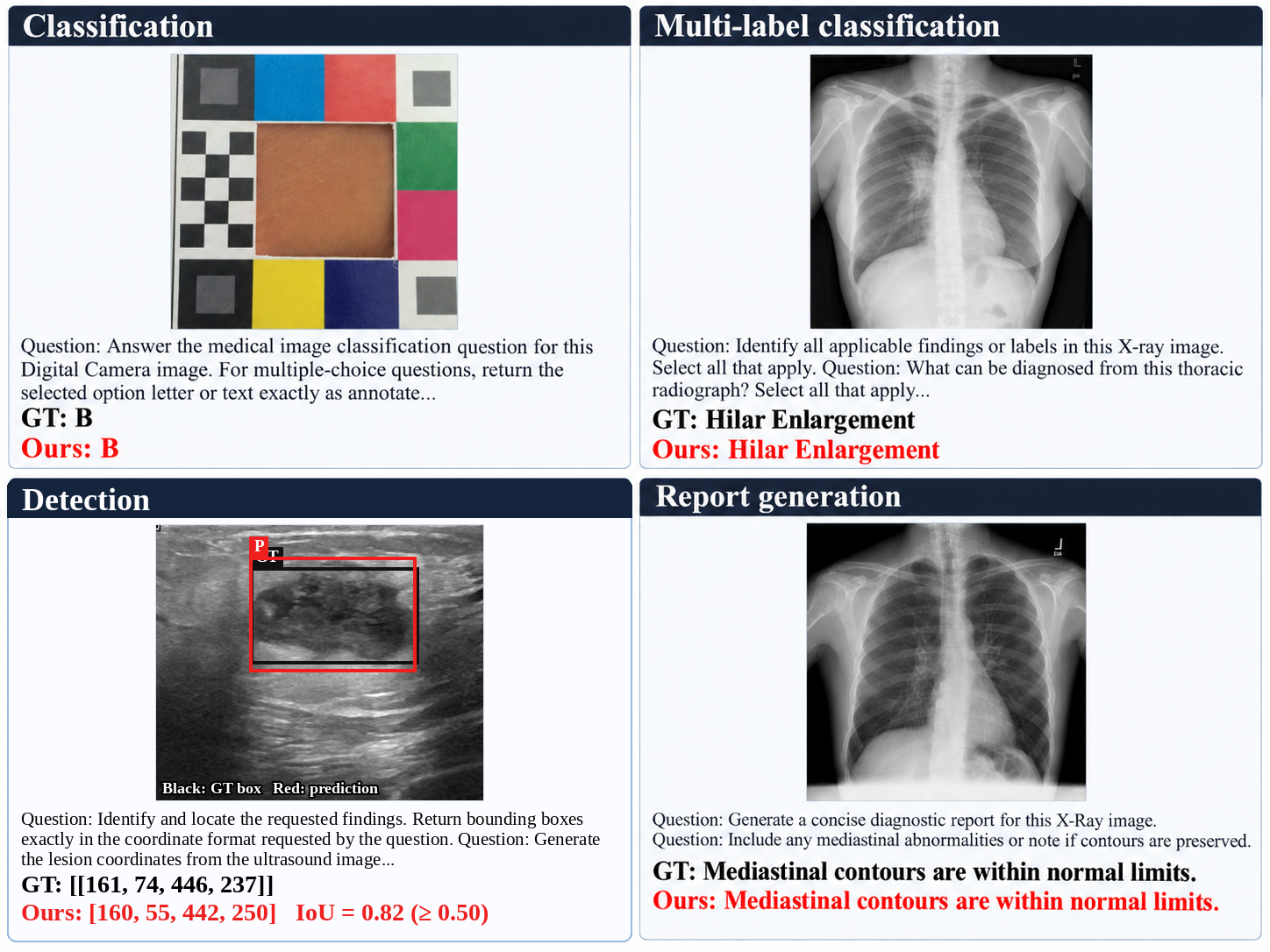}
  \caption{Qualitative comparison on the public-validation split. From left to right, examples cover classification,
  multi-label classification, detection, and report generation. Each example shows the input, task instruction, ground-truth, and our prediction.}
  \label{fig:qualitative}
\end{figure*}

Figure~\ref{fig:qualitative} presents representative examples from the public-validation split across classification, multi-label classification, detection, and report generation. For each example, we show the input image, ground-truth answer, and our prediction.

For classification, our task-specific expert generally follows the required answer format. The multi-label examples illustrate the completeness and precision of the predicted findings.
For report generation, the refined expert is able to produce reports that more closely match the content and descriptions of the reference reports in representative cases. In detection, the predicted bounding boxes show good agreement with the reference annotations in representative cases, indicating that the model can achieve accurate localization of the target findings.

\subsection{Results on final testing set}

{Table~\ref{tab:final_test} summarizes the performance of our final model on the official testing set. Specifically, the final evaluation is conducted on four tasks: detection, classification, multi-label classification, and numerical regression. These four tasks were selected because the challenge organizers only provided data for them and returned performance results exclusively for these tasks.

\begin{table}[!htbp]
\caption{Performance of the proposed method on the official final testing set.
$\uparrow$ indicates higher is better, while $\downarrow$ indicates lower is better.}
\label{tab:final_test}
\centering
\begin{tabular}{lcccc}
\hline
\textbf{Task}
& Detection
& Classification
& Multi-label Classification
& Regression
\\
\hline

\textbf{Metric}
& F1 $\uparrow$
& Balanced Acc. $\uparrow$
& {macro F1}  $\uparrow$
& MAE $\downarrow$
\\
\hline

\textbf{Ours}
& 0.7917
& 0.8507
& 0.4809
& 17.3954
\\

\hline
\end{tabular}
\end{table}

Our method achieves an F1 score of 0.7917 on detection and a balanced accuracy of 0.8507 on disease diagnosis classification. For multi-label classification, the model obtains a macro F1 score of 0.4809, while regression achieves an MAE of 17.3954. Comparing our validation results with the official test set reveals a performance gap, particularly in detection. We hypothesize that this discrepancy might be because the test set exhibits a distinct data distribution and domain shift compared to the validation set, potentially leaning toward specific target types or acquisition characteristics.

\subsection{Limitations and future work}
Fine-grained spatial perception remains the principal limitation of our method: although Stage 2 refinement improves detection, its F1 score remains below the strongest competing VLMs. Autoregressive coordinate generation is sensitive to small formatting and localization errors, which can invalidate an otherwise plausible box under an IoU threshold. 
The modality groups used for balanced resampling are benchmark-specific and may need to be redefined for a new dataset. Future work will investigate structured localization decoding, region-aware adapter designs, and more general sampling strategies for heterogeneous medical data.
Additionally, our two-stage Mixture-of-LoRA requires running both shared and task-specific adapters during inference, slightly increasing parameter footprint and latency. Future work will explore efficient adapter integration and merging strategies to mitigate this overhead.

\section{Conclusion}
We presented Two-Stage Mixture-of-LoRA, a parameter-efficient framework for multi-task medical vision-language learning. Stage 1 jointly trains a shared LoRA and task experts, while Stage 2 freezes the shared LoRA and all non-target experts to independently refine every task expert. Classification and regression then receive an additional continuation with within-task modality-balanced resampling. Experiments on FLARE 2026 Task 3 show that the combination improves classification, regression, detection, and counting relative to vanilla MedGemma-1.5-4B, while the ablations demonstrate that the optimal degree of specialization is task dependent. The framework offers a practical single-backbone route to multi-task medical image understanding and motivates future work on balanced adaptation and structured perception decoding.

\section*{Acknowledgements}
The authors declare that the proposed solution is fully automatic and does not require manual intervention. We thank all data owners and contributors for making the data publicly available and Codabench~\cite{xu2022codabench} for hosting the challenge platform.

\section*{Disclosure of Interests}
The authors declare no competing interests.
%
%
%
\bibliographystyle{splncs04}
\bibliography{paper}
\newpage
\begin{table}[!htbp]
\caption{Checklist Table. Please fill out this checklist table in the answer column.}
\centering
\begin{tabular}{ll}
\hline
Requirements                                                                                                                    & Answer        \\ \hline
A meaningful title                                                                                                              & Yes       \\ \hline
The number of authors ($\leq$6)                                                                                                             & 6        \\ \hline
Author affiliations and ORCID                                                                                           & Yes        \\ \hline
Corresponding author email is presented                                                                                                  & Yes        \\ \hline
Testing scores are presented in the abstract                                                                                 & Yes        \\ \hline
\begin{tabular}[c]{@{}l@{}}Introduction includes at least three parts: \\ background, related work, and motivation\end{tabular} & Yes        \\ \hline
A pipeline/network figure is provided                                                                                           & Figure 1 \\ \hline
The dataset and evaluation metric section are presented                                                                              & Page 8   \\ \hline
Environment setting table is provided                                                                                           & Table 1  \\ \hline
Training protocol table is provided                                                                                             & Table 3  \\ \hline
Ablation study                                                                                                                  & Page 13-15   \\ \hline
Efficiency evaluation results are provided                                                                                    & Table 3 \\ \hline
Limitation and future work are presented                                                                                        & Yes        \\ \hline
Reference format is consistent.  & Yes        \\ \hline
\end{tabular}
\end{table}
\end{document}